\def\year{2020}\relax
\documentclass[letterpaper]{article} 
\usepackage{aaai20}  
\usepackage{times}  
\usepackage{helvet} 
\usepackage{courier}  
\usepackage[hyphens]{url}  
\usepackage{graphicx} 
\usepackage{graphicx}  
\usepackage{bm}
\usepackage{amsfonts}
\usepackage[ruled,vlined]{algorithm2e}
\usepackage{multirow} 
\usepackage{mathrsfs}
\usepackage{subfigure}
\usepackage{color}
\usepackage{amssymb,amsmath}

\title{Mis-classified Vector Guided Softmax Loss for Face Recognition}
\author{Xiaobo Wang,\textsuperscript{\rm 1*} Shifeng Zhang,\textsuperscript{\rm 2\thanks{These authors contributed equally to this work.}} Shuo Wang,\textsuperscript{\rm 1} Tianyu Fu,\textsuperscript{\rm 1} Hailin Shi,\textsuperscript{\rm 1} Tao Mei\textsuperscript{\rm 1}\\
\textsuperscript{\rm 1}JD AI Research, Beijing, China\\
\textsuperscript{\rm 2}CBSR \& NLPR, Institute of Automation, Chinese Academy of Sciences, Beijing, China\\
wangxiaobo8@jd.com, shifeng.zhang@nlpr.ia.ac.cn, tmei@live.com
}
 
\begin{document}

\maketitle

\begin{abstract}
Face recognition has witnessed significant progress due to the advances of deep convolutional neural networks (CNNs), the central task of which is how to improve the feature discrimination. To this end, several margin-based (\textit{e.g.}, angular, additive and additive angular margins) softmax loss functions have been proposed to increase the feature margin between different classes. However, despite great achievements have been made, they mainly suffer from three issues: 1) Obviously, they ignore the importance of informative features mining for discriminative learning; 2) They encourage the feature margin only from the ground truth class, without realizing the discriminability from other non-ground truth classes; 3) The feature margin between different classes is set to be same and fixed, which may not adapt the situations very well. To cope with these issues, this paper develops a novel loss function, which adaptively emphasizes the mis-classified feature vectors to guide the discriminative feature learning. Thus we can address all the above issues and achieve more discriminative face features. To the best of our knowledge, this is the first attempt to inherit the advantages of feature margin and feature mining into a unified loss function. Experimental results on several benchmarks have demonstrated the effectiveness of our method over state-of-the-art alternatives. Our code is available at \url{http://www.cbsr.ia.ac.cn/users/xiaobowang/}.
\end{abstract}

\section{Introduction}
Face recognition is a fundamental and of great practice values task in the community of computer vision and pattern recognition. The task of face recognition contains two categories: face identification to classify a given face to a specific identity, and face verification to determine whether a pair of face images are of the same identity.
Though it has been extensively studied for decades \cite{USSDL,hu2017frankenstein,liu2019learning,hu2015face,wang2018support,chen2018virtual,wang2019co,liu2019adaptiveface,DeepID2+,shi2017cross}, there still exist a great many challenges for accurate face recognition, especially on large-scale test datasets at the very low false alarm rate (FAR), such as the MegaFace Challenge \cite{megaface_1,megaface_2} and the recent Trillion-Pairs Challenge \cite{deepglint}.

In recent years, the advanced face recognition models are usually built upon deep convolutional neural networks \cite{Attention56,Resnet,VGG} and the learned discriminative features play a significant role. To train deep models, the CNNs are generally equipped with classification loss functions \cite{Deepface,Center,SM-Softmax,SphereFace,EM-Softmax}, metric learning loss functions \cite{Contrastive,Facenet,Angular} or both \cite{DeepID2+,Center,zheng2018ring}. Metric learning loss functions such as contrastive loss \cite{Contrastive} or triplet loss \cite{Facenet} usually suffer from high computational cost. To avoid this problem, they require carefully designed sample mining strategies. But the performance is very sensitive to these strategies. So increasingly more researchers shift their attention to construct deep face recognition models by re-designing the classical classification loss functions.

Intuitively, face features are discriminative if their intra-class compactness and inter-class separability are well maximized. However, as pointed out by many recent studies \cite{Center,NormFace,SphereFace,AM-Softmax,EM-Softmax,Arc-Softmax}, the current prevailing classification loss function (\textit{i.e.}, Softmax loss) lacks the power of feature discrimination for deep face recognition. To address this issue, Wen \textit{et al.} \cite{Center} develop a center loss to learn centers for each identity to enhance the intra-class compactness. Wang \textit{et al.} \cite{NormFace} and Ranjan \textit{et al.} \cite{L2Constrain} propose to use a scale parameter to control the temperature of softmax loss, producing higher gradients to the well-separated samples to shrink the intra-class variance. Recently, several margin-based softmax loss functions \cite{L-softmax,SphereFace,cosface,AM-Softmax,Arc-Softmax} to increase the feature margin between different classes have also been proposed. Liu \textit{et al.} \cite{L-softmax,SphereFace}  introduce an angular margin (A-Softmax) between the ground truth class and other classes to encourage larger inter-class variance. However, it is usually unstable and the optimal parameters need carefully adjust for different settings. To enhance the stability of A-Softmax loss, Liang \textit{et al.} \cite{SM-Softmax} and Wang \textit{et al.} \cite{AM-Softmax,cosface} propose the additive margin (AM-Softmax) loss to stabilize the optimization. Deng \textit{et al.} \cite{Arc-Softmax}  develop an additive angular margin (Arc-Softmax) loss, which has a clear geometric interpretation.

Although the above approaches have achieved promising results, they mainly suffer from three shortcomings: 1) They obviously ignore the importance of informative features mining for discriminative learning. To address it, one may resort to the mining-based softmax loss functions. Shrivastava \textit{et al.} \cite{OHEM} design the hard mining strategy (HM-Softmax) to improve the feature discrimination by constructing mini-batches using high-loss examples. But the percentage of hard examples is empirically decided and the easy examples are completely discarded. In contrast, Lin \textit{et al.} \cite{Focal} design a relatively soft mining strategy, namely Focal loss (F-Softmax), to focus training on a sparse set of hard examples. However, the indication of hard examples is unclear. As a result, these two mining-based candidates usually fail to improve the performance. How to semantically select the hard examples is still an open problem. 2) They enlarge the feature margin only from the perspective of the ground truth class, which is partial and without realizing the discriminability from other non-ground truth classes. 3) Last but not at least, they enlarge the feature margin by using a same and fixed margin for all classes, which may not be appropriate and may not work very well in practice.

To overcome the aforementioned shortcomings, this paper tries to design a new loss function, which explicitly indicates the hard examples as mis-classified vectors and adaptively emphasizes on them to guide the discriminative feature learning. To sum up, the main contributions of this paper can be summarized as follows:
\begin{itemize}
\item{We propose a novel MV-Softmax loss, which explicitly indicates the hard examples and focuses on them to guide the discriminative feature learning. As a consequence, our new loss also absorbs the discrimiantibility from other non-ground truth classes as well as is with adaptive margins for different classes.}

\item{To the best of our knowledge, this is the first attempt to effectively inherit the merits of feature margin and feature mining techniques into a unified loss function. Moreover, We deeply analyze the relations and differences between our new loss  and the current margin-based and mining-based losses.}

\item{We conduct extensive experiments on the common benchmarks of LFW, CALFW, CPLFW, AgeDB, CFP, RFW, MegaFace and Trillion-Pairs, which have verified the superiority of our new approach over the baseline Softmax loss, the mining-based Softmax losses, the margin-based Softmax losses, and their naive fusions.}
\end{itemize}

\section{Preliminary Knowledge}
\noindent \textbf{Softmax}. Softmax loss is defined as the pipeline combination of last fully connected layer, softmax function and cross-entropy loss. In face recognition, the weights $\bm{w}_k$, (where $ k \in \{1,2,\dots,K\}$ and $K$ is the number of classes) and the feature $\bm{x}$ of the last fully connected layer are usually normalized and their magnitudes are replaced as a scale parameter $s$ \cite{NormFace,AM-Softmax,Arc-Softmax}. In consequence, given an input feature vector $\bm{x}$ with its corresponding ground truth label $y$, the softmax loss can be re-formulated as follows:
\begin{equation}\label{Softmax}
\begin{aligned}
\mathcal{L}_1 = - \log\frac{e^{s\cos(\theta_{\bm{w}_y,\bm{x}})}}{e^{s\cos(\theta_{\bm{w}_y,\bm{x}})}+\sum_{k\ne y}^Ke^{s\cos(\theta_{\bm{w}_k,\bm{x}})}},
\end{aligned}
\end{equation}
where $\cos(\theta_{\bm{w}_k,\bm{x}})=\bm{w}_k^T\bm{x}$ is the cosine similarity and $\theta_{\bm{w}_k,\bm{x}}$ is the angle between $\bm{w}_k$ and $\bm{x}$. As pointed out by a great many studies \cite{L-softmax,SphereFace,AM-Softmax,Arc-Softmax}, the learned features with softmax loss are prone to be separable, rather than to be discriminative for face recognition.

\noindent \textbf{Mining-based Softmax}. Hard example mining is becoming a common practice to effectively train deep CNNs. Its idea is to concentrate on informative examples, thus it usually results in more discriminative features. There are recent works that select hard examples based on loss value \cite{OHEM,Focal} to learn discriminative features. Generally, they can be summarized as:
\begin{equation}\label{Mining-Softmax}
\begin{aligned}
\mathcal{L}_2 = -g(p_y) \log\frac{e^{s\cos(\theta_{\bm{w}_y,\bm{x}})}}{e^{s\cos(\theta_{\bm{w}_y,\bm{x}})}+\sum_{k\ne y}^Ke^{s\cos(\theta_{\bm{w}_k,\bm{x}})}},
\end{aligned}
\end{equation}
where $p_y=\frac{e^{s\cos(\theta_{\bm{w}_y,\bm{x}})}}{e^{s\cos(\theta_{\bm{w}_y,\bm{x}})}+\sum_{k\ne y}^Ke^{s\cos(\theta_{\bm{w}_k,\bm{x}})}}$ is the predicted ground truth probability and $g(p_y)$ is an indicator function. Basically, for the soft mining method Focal loss \cite{Focal} (F-Softmax), $g(p_y)=(1-p_y)^\gamma$, $\gamma$ is a modulating factor. For the hard mining method HM-Softmax \cite{OHEM}, $g(p_y)=0$ when the sample is indicated as easy and $g(p_y)=1$ when the sample is hard.

\noindent \textbf{Margin-based Softmax}. To directly enhance the feature discrimination, several margin-based softmax loss functions \cite{SphereFace,EM-Softmax,AM-Softmax,Arc-Softmax} have been proposed in recent years. In summary, they can be defined as follows:
\begin{equation}\label{Margin-Softmax}
\begin{aligned}
\mathcal{L}_3 = - \log\frac{e^{sf(m,\theta_{\bm{w}_y,\bm{x}})}}{e^{sf(m,\theta_{\bm{w}_y,\bm{x}})}+\sum_{k\ne y}^Ke^{s\cos(\theta_{\bm{w}_k,\bm{x}})}},
\end{aligned}
\end{equation}
where $f(m,\theta_{\bm{w}_y,\bm{x}})$ is a carefully designed margin function. Basically,
$f(m_1,\theta_{\bm{w}_y,\bm{x}})= \cos(m_1\theta_{\bm{w}_y,\bm{x}})$ is the motivation of A-Softmax loss \cite{SphereFace}, where $m_1\ge1$ and is an integer. $f(m_2,\theta_{\bm{w}_y,\bm{x}})= \cos(\theta_{\bm{w}_y,\bm{x}})-m_2$ with $m_2 >0$ is the AM-Softmax loss \cite{AM-Softmax}. $f(m_3,\theta_{\bm{w}_y,\bm{x}})= \cos(\theta_{\bm{w}_y,\bm{x}}+m_3)$ with $m_3>0$ is the Arc-Softmax loss \cite{Arc-Softmax}. More generally, the margin function can be summarized into a combined version: $f(m,\theta_{\bm{w}_y,\bm{x}})=\cos(m_1\theta_{\bm{w}_y,\bm{x}} + m_3) - m_2$.

\section{Problem Formulation} \label{main}
To begin with, let us retrospect the formulation of margin-based softmax losses, \textit{i.e.}, Eq. (\ref{Margin-Softmax}), from which we can summarized that: 1) It ignores the importance of informative features mining for discriminative learning. 2) It only exploits the discriminability from the ground truth class $y$, \textit{i.e}, $f(m,\theta_{\bm{w}_y,\bm{x}})$, without be aware of the potential discriminability from other non-ground truth classes $k$, where $k \neq y$, $k\in \{1,2,\dots,K\} \backslash \{y\}$. 3) It simply uses a same and fixed margin $m_1$, $m_2$ or $m_3$ to enlarge the feature margin between different classes.

\subsection{Naive Mining-Margin Softmax Loss}
To solve the first shortcoming, one may resort to hard examples mining strategies \cite{OHEM,Focal}. The mining-based loss functions aim to focus training on the hard examples while the margin-based loss functions are to enlarge the feature margin between different classes. Therefore, these two branches are orthogonal and can seamlessly incorporate into each other, leading a naive motivation to directly integrate them as:
\begin{equation}\label{Mining-Margin-Softmax}
\begin{aligned}
\mathcal{L}_4 = -g(p_y) \log\frac{e^{sf(m,\theta_{\bm{w}_y,\bm{x}})}}{e^{sf(m,\theta_{\bm{w}_y,\bm{x}})}+\sum_{k\ne y}^Ke^{s\cos(\theta_{\bm{w}_k,\bm{x}})}}.
\end{aligned}
\end{equation}
The formulation Eq. (\ref{Mining-Margin-Softmax}) do involve informative features by the indicator function $g(p_y)$, but its improvement is limited in practice. The reason behind this may be, for the  HM-Softmax \cite{OHEM}, it explicitly indicates the hard examples, but it discards the easy ones. For the  F-Softmax \cite{Focal}, it uses all examples and empirically re-weights them by a modulating factor, but hard examples are unclear for training and without intuitive interpretation. This motivates us to design a more effective way to improve the performance.

\subsection{Mis-classified Vector Guided Softmax Loss}
\begin{figure}[t]
\centering
 \includegraphics[width=1.0\columnwidth]{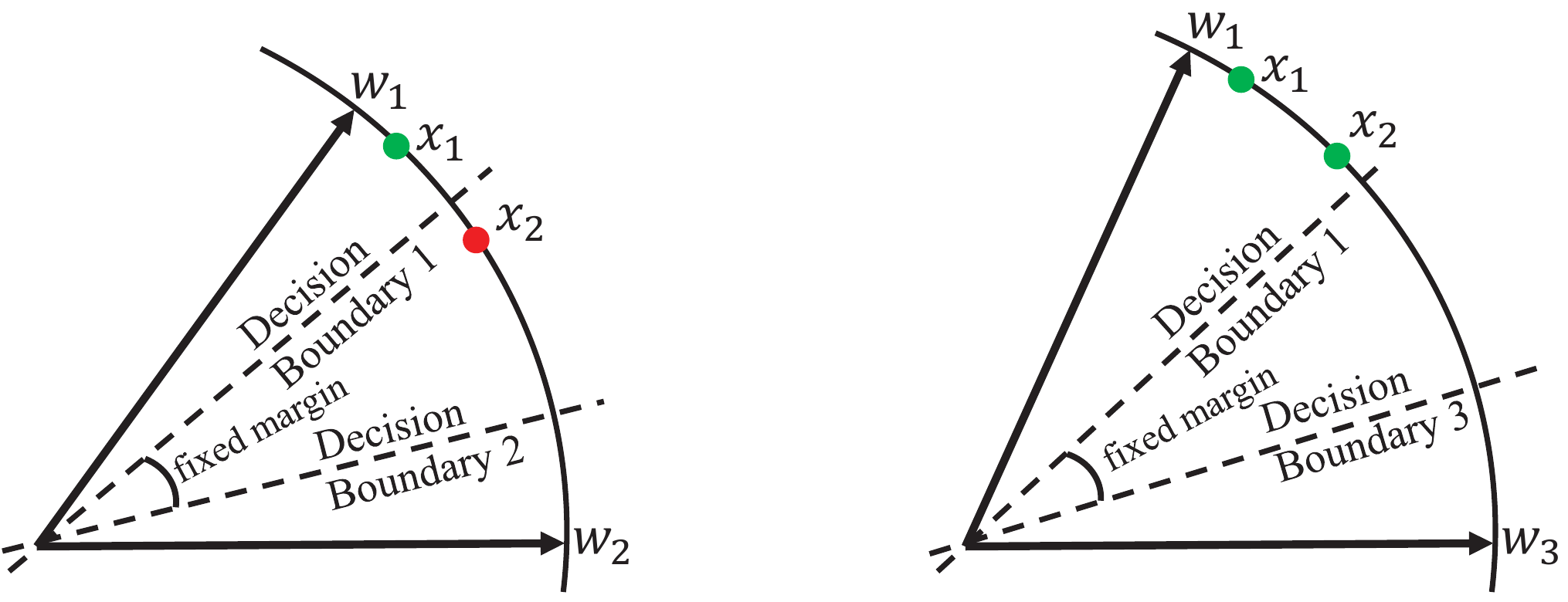}
  \caption{A geometrical interpretation of MV-Softmax from feature perspective. Samples $\bm{x}_1$ and $\bm{x}_2$ are both from class 1. The mis-classified vectors (red dots) are those who are mis-classified by a specific classifier (\textit{e.g.}, $\bm{w}_2$).}
\label{fig:sv-mining}
\end{figure}

Intuition says that considering the well-separated feature vectors has little effect on the learning problem. That means the mis-classified feature vectors are more crucial to enhance feature discriminability. To this end, we alternatively introduce a more elegant way to focus training on the truly informative features (\textit{i.e.}, mis-classified vectors). Specifically, based on the margin-based softmax loss functions, we define a binary indicator $I_k$ to adaptively indicate whether a sample (feature) is mis-classified by a specific classifier $\bm{w}_k$ (where $k\neq y$) in the current stage:
\begin{equation}\label{weight}
\begin{aligned}
\ \  {I}_k =
 \left \{
   \begin{aligned}
  & 0,  \ \ f(m,\theta_{\bm{w}_y,x})-\cos(\theta_{\bm{w}_k,x})\geq 0 \\
  & 1, \ \ f(m,\theta_{\bm{w}_y,x})-\cos(\theta_{\bm{w}_k,x})< 0\\
   \end{aligned}
\right. .
\end{aligned}
\end{equation}
From the definition Eq. (\ref{weight}), we can see that if a sample (feature) is mis-classified, \textit{i.e.}, $f(m,\theta_{\bm{w}_y,\bm{x}})-\cos(\theta_{\bm{w}_k,\bm{x}})<0$ (\textit{e.g.}, in the left sub-figure of Figure \ref{fig:sv-mining}, the feature $\bm{x}_2$ belongs to class 1, but it is mis-classified by the classifier $\bm{w}_2$, \textit{i.e.}, $f(m,\theta_{\bm{w}_1,\bm{x}_2})-\cos(\theta_{\bm{w}_2,\bm{x}_2})<0$), it will be emphasized temporarily. In this way, the hard examples are explicitly indicated and we mainly focus on them for discriminative training. Consequently, we formulate our Mis-classified Vector guided Softmax (\textbf{MV-Softmax}) loss as follows:
\begin{equation}\label{SV-Softmax}
\noindent \mathcal{L}_5 = -\log  \frac{e^{sf(m,\theta_{\bm{w}_y,\bm{x}})}}{e^{sf(m,\theta_{\bm{w}_y,\bm{x}})}+\sum_{k\neq y}^K h(t,\theta_{\bm{w}_k,\bm{x}},I_k)e^{s\cos(\theta_{\bm{w}_k,\bm{x}})}},
\end{equation}
where $h(t,\theta_{\bm{w}_k,\bm{x}},I_k) \geq 1$ is a re-weighted function to emphasize the indicated mis-classified vectors. Here we give two candidates, one is with fixed weights for all mis-classified classes:
\begin{equation} \label{indicator1}
 h(t,\theta_{\bm{w}_k,\bm{x}},I_k)=e^{stI_k},
\end{equation}
and the other one is an adaptive formulation:
\begin{equation} \label{indicator2}
 h(t,\theta_{\bm{w}_k,\bm{x}},I_k)=e^{st(\cos(\theta_{\bm{w}_k,\bm{x}})+1)I_k}.
\end{equation}
where $t\geq 0$ is a preset hyperparameter. Obviously, when $t=0$, the designed MV-Softmax loss Eq. (\ref{SV-Softmax}) becomes identical to the original margin-based softmax losses Eq. (\ref{Margin-Softmax}).

\subsubsection{Comparision to Mining-based Softmax Losses.}
To illustrate the advantages of our MV-Softmax loss over the traditional mining-based loss functions (\textit{e.g.}, HM-Softmax \cite{OHEM} and F-Softmax \cite{Focal}), Figure \ref{fig:sv-mining} gives a toy example. Assume that we have two samples (features) $\bm{x}_1$ and $\bm{x}_2$, both of them are from class 1, where $\bm{x}_1$ is well-classified while $\bm{x}_2$ is not. The HM-Softmax empirically indicates the hard samples and discards the easy sample $\bm{x}_1$ to use the hard one $\bm{x}_2$ for training. The F-Softmax does not explicitly indicate the hard samples, but it re-weights all the samples, making the harder one $\bm{x}_2$ to have relatively larger loss value. These two strategies are directly from the loss viewpoint and the selection of hard examples is without semantic guidance. Our MV-Softmax loss Eq. (\ref{SV-Softmax}) is from a different way. Firstly, we semantically indicates the hard examples (mis-classified vectors) according to the decision boundary. The hardness of previous methods is defined as a global relationship between feature (sample) and feature (sample). While our hardness is a local relationship between feature and classifier, which is more  consistent with discriminative feature learning. Then, we emphasize these hard examples from probability viewpoint. Specifically, because the cross-entropy loss $-log(p)$ is a monotonically decreasing function, reducing the probability $p$ (the reason is that $h(t,\theta_{\bm{w}_k,\bm{x}},I_k) \geq 1$, see Eqs. (\ref{indicator1}) and (\ref{indicator2})) of the mis-classified vector $\bm{x}_2$, will increase its importance for training. In summary, we can claim that our mis-classified vector guided mining strategy, is more superior for discriminative feature learning than previous ones.

\subsubsection{Comparision to Margin-based Softmax Losses.}
Similarly, assume that we have a sample $\bm{x}_2$ from class 1, and it is not well-classified, (\textit{e.g.}, the red dot in Figure \ref{fig:sv-mining}). The original softmax loss aims to make $\bm{w}_1^T\bm{x}_2>\bm{w}_2^T\bm{x}_2 \Longleftrightarrow \cos(\theta_1) > \cos(\theta_2)$ and $\bm{w}_1^T\bm{x}_2>\bm{w}_3^T\bm{x}_2 \Longleftrightarrow \cos(\theta_1) > \cos(\theta_3)$. To make these objectives more rigorous, margin-based loss functions introduce a margin function $f(m,\theta_1)=\cos(m_1\theta_1+m_3)-m_2$ from the perspective of ground truth class (\textit{i.e.}, $\theta_1$) \cite{SphereFace,AM-Softmax,Arc-Softmax}:
\begin{equation}
\begin{aligned}
 & \cos(\theta_1)\geq f(m,\theta_1)>\cos(\theta_2) \\
 & \cos(\theta_1)\geq f(m,\theta_1)>\cos(\theta_3),
 \end{aligned}
\end{equation}
wherein $f(m, \theta_1)$ is with a same and fixed margin for different classes and ignores the potential discriminability from other non-ground truth classes (\textit{e.g.}, $\theta_2$ and $\theta_3$). To solve these issues, our MV-Softmax loss tries to further enlarge the feature margin from the perspective of other non-ground truth classes. Specifically, we have introduced a margin function $h^*(t,\theta_2)$ for the mis-classified feature $\bm{x}_2$:
\begin{equation}
\begin{aligned}
& \cos(\theta_1)\geq f(m,\theta_1)>h^*(t,\theta_2)\geq \cos(\theta_2) \\
& \cos(\theta_1)\geq f(m,\theta_1)\geq \cos(\theta_3),
\end{aligned}
\end{equation}
where $h^*(t,\theta_2)=\log[h(t,\theta_2)e^{\cos(\theta_2)}]=\cos(\theta_2)+t$ or $(t+1)\cos(\theta_2)+t$. For the case $\theta_3$, because $\bm{x}_2$ is well-classified by the classifier $\bm{w}_3$, we do not need to give any additional enforcement to further enlarge its margin. Moreover, our MV-Softmax losses have also set adaptive margins for different classes. Taking MV-AM-Softmax (\textit{i.e.}, $f(m, \theta_y) = \cos(\theta_y)-m$) as an example, for the mis-classified classes, the margin is $m+t$ or $m+t\cos(\theta_2)+t$. While for the well-classified classes, the margin is $m$. On account of these, our MV-Softmax losses have addressed the second and third shortcomings.

According to the above discussions,  we conclude that our new loss has inherited the merits of feature margin and feature mining into a unified loss function, thus it is expected to achieve more discriminative features for face recognition.

\subsection{Optimization}
In this section, we show that our MV-Softmax loss Eq.  (\ref{SV-Softmax}) is trainable and can be easily optimized by the typical stochastic gradient descent (SGD). The difference between the previous margin-based softmax losses and the proposed MV-Softmax loss lies in the last fully connected layer
$\bm{v}=[v_1,v_2,\dots,v_K]^T=[\cos(\theta_{\bm{w}_1,\bm{x}}),\cos(\theta_{\bm{w}_2,\bm{x}}),\dots,\cos(\theta_{\bm{w}_K,\bm{x}})]^T$. For the forward propagation, when $k=y$, it is the same as the original margin-based softmax loss (\textit{i.e.}, $v_y=\cos(m_1\theta_{\bm{w}_y,\bm{x}}+m_3)-m_2$). When $k\neq y$, it has two cases, if the feature vector is well-classified for a specific classifier, it is the same as the original softmax (\textit{i.e.},  $v_k=\cos(\theta_{\bm{w}_k,\bm{x}})$). Otherwise, it will be re-computed with a fixed weight $\cos(\theta_{\bm{w}_k,\bm{x}})+t$ or an adaptive weight  $(t+1)\cos(\theta_{\bm{w}_k,\bm{x}})+t$. The whole scheme of our method is summarized in Algorithm \ref{algorithm}.

\begin{algorithm}[t] \label{algorithm}
	\SetAlgoLined \caption{\small MV-Softmax}
	\begin{small}
		\KwIn{Training set $\mathcal{S}$; The hyper-parameter $t$; Training epochs $\tau$.}
		
		\textbf{Initialization}: $\alpha=1$; Randomly initialize the parameter $\bm{\Theta}$ in convolution layers and $\bm{W}$ in the last fully connected layer.
		
		\While{$\alpha \leq \tau$}{

          Shuffle the training set $\mathcal{S}$ and fetch mini-batch $\mathcal{S}_n$;

          \textbf{Forward}: According to the indication of hard examples Eq. (\ref{weight}), we compute the MV-Softmax loss by Eq. (\ref{SV-Softmax});

          \textbf{Backward}: Update the parameters $\bm{W}$ and $\bm{\Theta}$ by Stochastic Gradient Descent (SGD);
		}
		\KwOut{Parameters $\bm{\Theta}$ and $\bm{W}$.}
	\end{small}
\end{algorithm}

\section{Experiments}
\begin{table}[t]
\caption{Face datasets for training and test. "(P)" and "(G)" refer to the probe and gallery set, respectively.}\smallskip
\centering
\resizebox{1.0\columnwidth}{!}{
\smallskip\begin{tabular}{|c|c|c|c|}
\hline
& Datasets & Identities   & Images     \\
\hline\hline
 Training & MS-Celeb-1M-v1c-R & 72,690 & 3.28M	 \\
 \hline
 \multirow{8}{*}{Test}
 & LFW & 5,749 & 13,233	 \\
 & CALFW  & 5,749 & 12,174	 \\
 & CPLFW  & 5,749 & 11,652 \\
 & AgeDB  & 568 & 16,488	 \\
 & CFP  & 500 & 7,000 \\
 & RFW  & 11,430 & 40,607 \\
 & MegaFace  & 530(P) & 1M(G) \\
 & Trillion-Pairs  & 5,749(P) & 1.58M(G)\\
 \hline
\end{tabular}}
\label{data-statics}
\end{table}

\subsection{Datasets}
\noindent \textbf{Training Data}. The original MS-Celeb-1M dataset \cite{Msceleb} contains about 100K identities with 10M images. However, it consists of a great many noisy faces. Fortunately, the trillion-pairs consortium \cite{deepglint} has made their efforts to get a high-quality version MS-Celeb-1M-v1c, which is well-cleaned for training.

\noindent \textbf{Test Data}. We use eight face recognition benchmarks, including LFW \cite{LFW}, CALFW \cite{CALFW}, CPLFW \cite{CPLFW}, AgeDB \cite{AgeDB}, CFP \cite{CFP}, RFW \cite{RFW}, MegaFace \cite{megaface_1,megaface_2} and Trillion-Pairs \cite{deepglint}, as the test data. For more details about the test datasets, please see their references.

\noindent \textbf{Dataset Overlap Removal}. In face recognition, it is very important to perform open-set evaluation, \textit{i.e.}, there should be no overlapping identities between training set and test set. To this end, we need to carefully remove the overlapped identities between the employed training dataset (\textit{i.e.}, MS-Celeb-1M-v1c) and the test datasets (including LFW, CALFW, CPLFW, AgeDB, CFP, RFW and MegaFace)\footnote{For the Trillion-Pairs test set, we can not remove the potential overlaps because its ground truth label (name) is unreleased.}. For the overlap identities removal tool, we use the publicly available script provided by \cite{AM-Softmax} to check whether if two names are of the same person. As a consequence, we remove 14,186 identities from the training set MS-Celeb-1M-v1c. For clarity, we donate the refined training dataset as MS-Celeb-1M-v1c-R. Important statistics of all the involved datasets are summarized in Table \ref{data-statics}. To be rigorous, all the experiments in this paper are based on the refined training set MS-Celeb-1M-v1c-R. To encourage more researchers to abide by the open-set protocol, the overlapping lists and the refined dataset MS-Celeb-1M-v1c-R are publicly available.

\begin{table}[t]
\caption{Verification performance (\%) of our MV-Softmax loss functions with different hyper-parameter $t$. 'f' and 'a' donate the fixed re-weight function Eq. (\ref{indicator1}) and the adaptive one Eq. (\ref{indicator2}), respectively.}\smallskip
\centering
\resizebox{1.0\columnwidth}{!}{
\smallskip
\begin{tabular}{|c|c|c|c|c| }
\hline
\multirow{2}{*}{Method} & BLUFR & \multirow{2}{*}{CALFW} & \multirow{2}{*}{AgeDB}  \\
  & 1e-5 &  &       \\
\hline\hline
MV-Arc-Softmax-f (0.15)         & 94.60 & \textbf{95.54} & 98.05  \\
MV-Arc-Softmax-f (0.2)          & \textbf{95.18} & 95.46 & \textbf{98.11}  \\
MV-Arc-Softmax-f (0.25)         & 94.04 & 95.51 & 98.08  \\
\hline\hline
MV-Arc-Softmax-a (0.25)         & 94.15 & 95.33 & 97.86   \\
MV-Arc-Softmax-a (0.3)          & \textbf{95.50} & 95.46 & \textbf{98.06}  \\
MV-Arc-Softmax-a (0.35)         & 95.08 & \textbf{95.50} & 97.90   \\
\hline\hline
MV-AM-Softmax-f (0.2)           & 94.81 & 95.29 & 98.01  \\
MV-AM-Softmax-f (0.25)          & \textbf{95.74} & \textbf{95.45} & \textbf{98.05}  \\
MV-AM-Softmax-f (0.3)           & 95.07 & 95.41 & 98.00  \\
\hline\hline
MV-AM-Softmax-a (0.15)          & 94.09 & 95.41 & \textbf{98.13}  \\
MV-AM-Softmax-a (0.2)           & \textbf{96.27} & \textbf{95.63} & 98.00  \\
MV-AM-Softmax-a (0.25)          & 94.29 & 95.51 & 97.96  \\
\hline
\end{tabular}}
\label{Effect}
\end{table}

\begin{table*}[t]
\caption{Verification performance (\%) of different loss functions on the test sets LFW, CALFW, CPLFW, AgeDB and CFP.}\smallskip
\centering
\resizebox{2.0\columnwidth}{!}{
\smallskip
\begin{tabular}{|c|c|c|c|c|c|c|c|c|c| }
\hline
& \multirow{2}{*}{Method}  & \multirow{2}{*}{LFW} & \multicolumn{3}{|c|}{BLUFR} & \multirow{2}{*}{CALFW} & \multirow{2}{*}{CPLFW} & \multirow{2}{*}{AgeDB} & \multirow{2}{*}{CFP}     \\
\cline{4-6}
&         &  & 1e-3      & 1e-4      & 1e-5 & & & &       \\

\hline\hline
Baseline & Softmax                       & 99.59 & 99.29 & 99.11 & 91.74 & 94.66 & 87.76 & 97.01 & 94.04 \\
\hline
\multirow{2}{*}{Mining-based}
& F-Softmax         & 99.65 & 99.24 & 98.72 & 91.19  & 93.83 & 86.35 & 96.51 & 93.20  \\
& HM-Softmax        & 99.65 & 99.30 & 99.11 & 92.03  & 94.69 & 87.56 & 97.05 & 94.12  \\
\hline
\multirow{3}{*}{Margin-based}
& A-Softmax    & 99.65 & 99.30 & 99.12 & 92.77 & 94.55 & 87.85 & 97.16 & 94.22   \\
& Arc-Softmax  & 99.76 & 99.33 & 99.30 & 93.75 & 95.44 & 88.78 & 98.00 & 95.28 \\
& AM-Softmax   & 99.71 & 99.33 & 99.31 & 93.68 & 95.58 & 89.60 & 98.03 & 95.68   \\
\hline
\multirow{4}{*}{Naive-fused}
& F-Arc-Softmax                    & 99.71 & 99.33 & 99.29 & 94.51 & 95.48 & 88.85 & 98.10 & 95.62 \\
& F-AM-Softmax  & 99.73 & 99.33 & 99.30 & 92.81 & 95.58 & 89.60 & \textbf{98.20} & 95.47   \\
& HM-Arc-Softmax               & 99.75 & 99.33 & 99.29 & 93.53 & 95.36 & 89.16 & 97.86 & 95.22 \\
& HM-AM-Softmax                & 99.76 & 99.33 & 99.30 & 96.09 & 95.45 & 89.56 & 98.05 & 95.37  \\
\hline
\multirow{4}{*}{Ours}
& MV-Arc-Softmax-f (0.2)       & 99.78 & \textbf{99.34} & 99.30 & 95.18 & 95.46 & 89.30 & 98.11 & 95.21 \\
& MV-Arc-Softmax-a (0.3)       & 99.76 & 99.33 & 99.30 & 95.50 & 95.46 & 89.41 & 98.06 & 95.45 \\
\cline{2-10}
& MV-AM-Softmax-f (0.25)       & 99.79 & 99.33 & \textbf{99.31} & 95.74 & 95.45 & \textbf{89.69} & 98.05 & \textbf{95.70} \\
& MV-AM-Softmax-a (0.2)        & \textbf{99.79} & 99.33 & 99.30 & \textbf{96.27} & \textbf{95.63} & 89.19 & 98.00 & 95.30 \\
\hline
\end{tabular}}
\label{LFW1}
\end{table*}

\subsection{Experimental Settings}
\noindent \textbf{Data Processing}.
We detect the faces by adopting the FaceBoxes detector \cite{facebox,zhang2019faceboxes} and localize five landmarks (two eyes, nose tip and two mouth corners) through a simple 6-layer CNN \cite{feng2017wing,liu2019high}. The detected faces are cropped and resized to 144$\times$144, and each pixel (ranged between [0,255]) in RGB images is normalized by subtracting 127.5 and divided by 128. For all the training faces, they are horizontally flipped with probability 0.5 for data augmentation.

\noindent \textbf{CNN Architecture}. In face recognition, there are many kinds of network architectures \cite{SphereFace,AM-Softmax,wang2018devil}. To be fair, the CNN architecture should be the same to test different loss functions. As suggested by the work \cite{wang2018devil}, we use the AttentionNet \cite{Attention56} to achieve a good balance between computation and accuracy. Moreover, inspired by the work \cite{Arc-Softmax}, we integrate the IRSE module into the AttentionNet and rename the developed architecture as AttentionNet-IRSE. For the depth stages of AttentionNet-IRSE, we set [1, 1, 1] as our baseline architecture. The output of AttentionNet-IRSE gets a 512-dimension feature.

\noindent \textbf{Training}. All the CNN models are trained with stochastic gradient descent (SGD) algorithm and are trained from scratch, with the batch size of 32 on 4 P40 or 4 V100 GPUs parallelly, total batch size 128. The weight decay is set to 0.0005 and the momentum is 0.9. The learning rate is initially 0.1 and divided by 10 at 4, 8, 10 epochs, and we finish the training process at 12 epoch. All experiments in this paper are implemented by Pytorch library.

\noindent \textbf{Test}. At test stage, only the original image features are employed to compose the face representation. All the reported results in this paper are evaluated by a single model, without model ensemble or other fusion strategies.

For the evaluation metric, the cosine similarity is utilized. We follow the unrestricted with labelled outside data protocol \cite{LFW} to report the performance on LFW, CALFW, CPLFW, AgeDB, CFP and RFW. Moreover, we also report the BLUFR protocol \cite{blufr} on the test set LFW. On Megaface and Trillion-Pairs Challenge, face identification and verification are conducted by ranking and thresholding the scores. Specifically, for face identification, the Cumulative Match Characteristics (CMC) curves are adopted to evaluate the Rank-1 accuracy. For face verification, the Receiver Operating Characteristic (ROC) curves are adopted. The true positive rate (TPR) at low false acceptance rate (FAR) is emphasized since in real applications false acceptance gives higher risks than false rejection.

For the compared methods, we compare our method with the baseline Softmax loss (\textbf{Softmax}) and the recently proposed state-of-the-arts, including 2 mining-based softmax losses (\textit{i.e.}, \textbf{F-Softmax} and \textbf{HM-Softmax}), 3 margin-based softmax losses (\textbf{A-Softmax}, \textbf{Arc-Softmax} and \textbf{AM-Softmax}) and their 4 naive fusions (\textbf{F-Arc-Softmax}, \textbf{F-AM-Softmax}, \textbf{HM-Arc-Softmax}  and \textbf{HM-AM-Softmax}). For all the competitors, their source codes can be downloaded from the github or from authors' webpages. The corresponding parameters of each competitors are mainly determined according to their paper's suggestions. Specifically, for HM-Softmax \cite{OHEM}, we save 90\% high-loss samples in each mini-batch for training.  For F-Softmax, it is with the parameter $\gamma=2.0$. For A-Softmax, the margin parameter is set as $m_1=3$. While for AM-Softmax and Arc-Softmax, the margin parameters are set as $m_2=0.35$ and $m_3=0.5$, respectively. The scale parameter $s$ has already been discussed sufficiently in previous works \cite{AM-Softmax,cosface}. In this paper, we empirically fixed it to 32 for all the methods.

\subsection{Exploratory Experiments}
\noindent \textbf{Effect of parameter $t$}. Since the hyper-parameter $t$ in the re-weighted function Eqs. (\ref{indicator1}) and (\ref{indicator2}) plays an important role in the developed MV-Softmax loss, we mainly explore to search its possible best value in this part. In Table \ref{Effect}, we list the performance of our proposed MV-Softmax loss function with $t$ varies from different ranges. 'f' and 'a' donate the fixed re-weight function Eq. (\ref{indicator1}) and the adaptive one Eq. (\ref{indicator2}), respectively. From the numbers, we can summarize that our MV-Softmax loss is insensitive to the hyper-parameter $t$ in a certain range. Moreover, according to this study, we empirically set $t=0.2$ for MV-Arc-Softmax-f, $t=0.3$ for MV-Arc-Softmax-a, $t=0.25$ for MV-AM-Softmax-f and $t=0.2$ for MV-AM-Softmax-a in the subsequent experiments.

\begin{figure}[t]
\centering
\includegraphics[width=1.0\columnwidth]{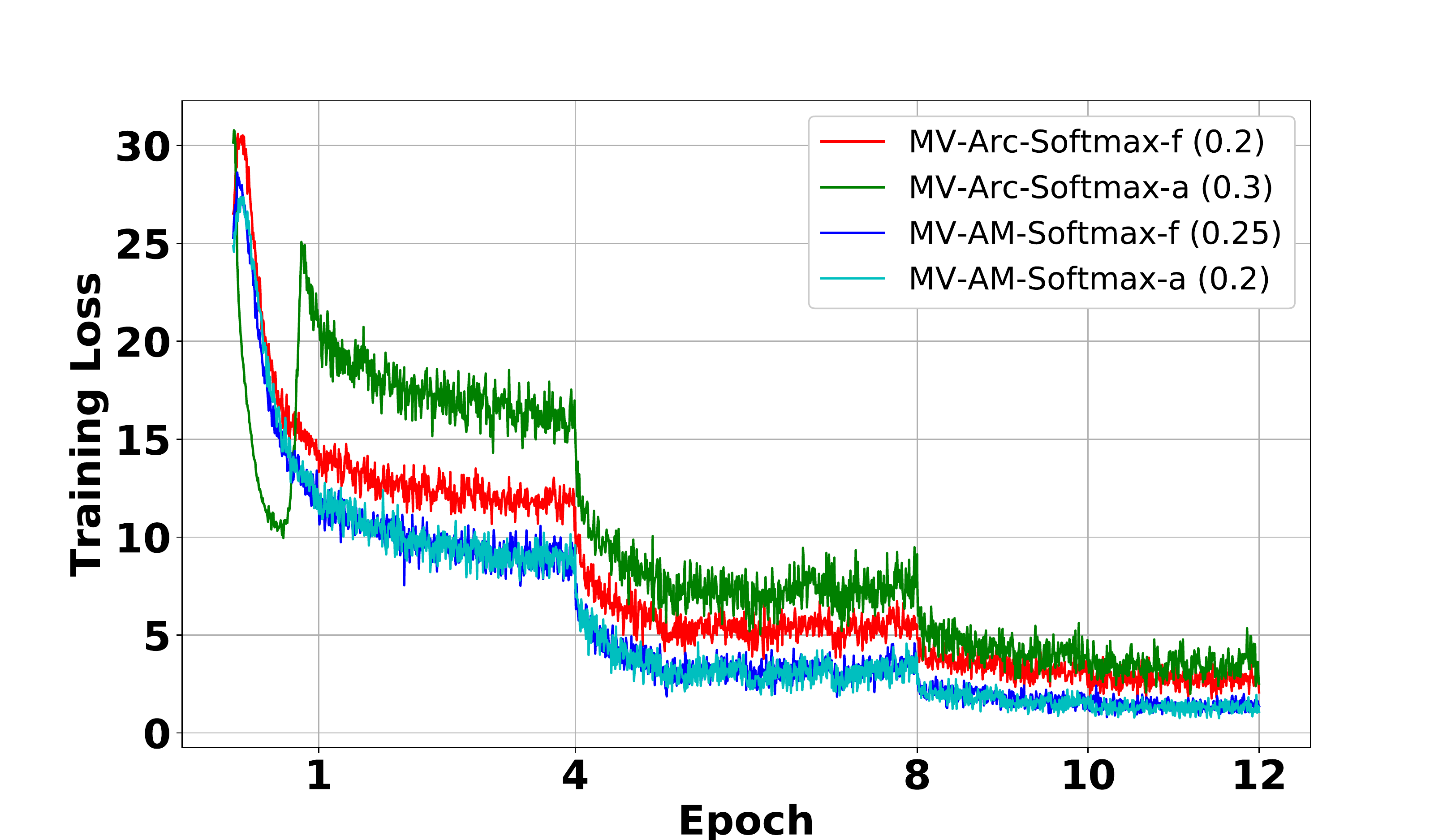}
   \caption{Convergence of MV-Softmax. From the curves, we can see that our MV-Softmax loss functions have a good behavior of convergence.}
\label{fig:convergence}
\end{figure}

\noindent \textbf{Convergence of MV-Softmax}. Although the convergence of our method is not easy to be theoretically analyzed, it would be intuitive to see its empirical behavior. Here, we give the loss changes as the number of epochs increases. From the curves in Figure \ref{fig:convergence}, it can be observed that our method has a good behavior of convergence.

\subsection{Results on LFW, CALFW, CPLFW, AgeDB, CFP}
Table \ref{LFW1} provides the quantitative results of all the competitors on LFW, CALFW, CPLFW, AgeDB and CFP. The bold number in each column represents the best result. For the LFW accuracy and its BLUFR protocol with different false alarm rates  (\textit{e.g.}, 1e-3, 1e-4, 1e-5), it is well-known that these protocols are typical and easy for face recognition. For instance, at LFW accuracy and TPR@FAR=1e-3 and 1e-4, almost all the competitors can achieve 99\% performance. So the improvement of our MV-Softmax losses is not quite large. For the BLUFR with TPR@FAR=1e-5, we can see that the naive fusion HM-AM-Softmax outperforms the baseline Softmax, the simple mining-based losses and the margin-based ones. Despite this, our MV-AM-Softmax still achieves about 0.2\% improvement. On CALFW, CPLFW, AgeDB and CFP test sets, we also observe that our MV-Softmax losses are better than the state-of-the-art alternatives in most of cases. Nevertheless, we can see that the improvements of our method in these test sets are not by a large margin. The reason is that the test protocol is relatively easy and the performance of all the methods on these test sets are near saturation. So there is an urgent need to test the performance of all the competitors on new test sets or test with more complicated protocols.

\begin{table}[t]
\caption{Verification performance (\%) of different loss functions on the test set RFW. }\smallskip
\centering
\resizebox{1.0\columnwidth}{!}{
\smallskip
\begin{tabular}{|c|c|c|c|c| }
\hline
\multirow{2}{*}{Method} & \multicolumn{4}{|c|}{RFW}  \\
\cline{2-5}
& Caucasian & Indian & Asian  & African  \\
\hline\hline
Softmax & 98.33 & 93.33 & 93.16 & 91.33 \\
\hline
F-Softmax & 97.50 & 90.30 & 91.16 & 88.33 \\
HM-Softmax & 98.66 & 93.49 & 92.83 & 90.50 \\
\hline
A-Softmax   & 98.83 & 94.33 & 93.33 & 91.33 \\
Arc-Softmax & 98.83 & 96.16 & 93.66 & 95.00 \\
AM-Softmax  & 99.16 & 96.16 & 94.46 & 95.83 \\
\hline
F-Arc-Softmax                   & 98.99 & 95.83 & 94.16 & 95.50   \\
F-AM-Softmax                    & 99.16 & 96.66 & 93.66 & 95.00   \\
HM-Arc-Softmax                  & 98.66 & 94.33 & 94.16 & 96.66   \\
HM-AM-Softmax                   & 99.16 & 94.66 & 93.33 & 96.00   \\
\hline
MV-Arc-Softmax-f              & 98.66 & \textbf{96.83} & 94.50 & 96.50  \\
MV-Arc-Softmax-a              & 98.00 & 94.66 & 94.83 & 95.99  \\
MV-AM-Softmax-f               & 99.00 & 94.99 & 94.83 & \textbf{96.66}  \\
MV-AM-Softmax-a               & \textbf{99.33} & 95.83 & \textbf{95.66} & 95.83 \\
 \hline
\end{tabular}}
\label{RFW-1}
\end{table}

\subsection{Results on RFW}
Firstly, we evaluate all the competitors on the recent proposed new test set RFW \cite{RFW}. RFW is a face recognition benchmark for measuring racial bias, which consists of four test subsets, namely Caucasian, Indian, Asian and African. Tables \ref{RFW-1} displays the performance comparison of all the involved methods. From the values, we can conclude that the results on the four subsets exhibit the same trends, \textit{i.e.}, the margin-based losses are better than the baseline Softmax loss and the mining-based losses. The improvement by simply combining the margin-based and mining-based losses is limited. Our mis-classified guided ones, which explicitly emphasize on the mis-classified feature vectors for training, are more consistent with the discriminative feature learning. Therefore, they inherently absorb the merits of feature margin and feature mining into a unified loss function. They usually achieve more discriminative face features and can get higher performance than previous alternatives.

\begin{table}[t]
\caption{Performance (\%) of different loss functions on MegaFace and Trillion-Pairs Challenge.}\smallskip
\centering
\resizebox{1.0\columnwidth}{!}{
\smallskip
\begin{tabular}{|c|c|c|c|c| }
\hline
\multirow{2}{*}{Method} & \multicolumn{2}{|c|}{MegaFace} &\multicolumn{2}{|c|}{Trillion-Pairs} \\
\cline{2-5}
& Id.  & Veri. & Id.  & Veri.  \\
\hline\hline
Softmax     & 93.94 & 94.76 & 60.06 & 59.00\\
\hline
F-Softmax   & 91.60 & 93.06 & 51.14 & 48.32\\
HM-Softmax  & 93.95 & 95.53 & 61.34 & 60.07 \\
\hline
A-Softmax   & 94.18 & 95.26 & 60.34 & 59.01 \\
Arc-Softmax & 97.28 & 97.58 & 70.80 & 68.12 \\
AM-Softmax  & 97.69 & 97.82 & 74.00 & 71.57 \\
\hline
F-Arc-Softmax                   & 97.51 & 97.81 & 70.65 & 69.06   \\
F-AM-Softmax                    & 95.75 & 97.75 & 73.82 & 72.18 \\
HM-Arc-Softmax                  & 97.43 & 97.56 & 70.08 & 68.16    \\
HM-AM-Softmax                   & 97.48 & 97.64 & 73.89 & 71.63    \\
\hline
MV-Arc-Softmax-f              & 97.52 & 98.01 & 73.90 & 71.28  \\
MV-Arc-Softmax-a              & 97.74 & 97.62 & 75.44 & 74.69  \\
MV-AM-Softmax-f               & 97.95 & 97.85 & 75.92 & 74.45  \\
MV-AM-Softmax-a               & \textbf{98.00} & \textbf{98.31} & \textbf{76.94} & \textbf{75.93} \\
 \hline
\end{tabular}}
\label{MegaFace}
\end{table}

\subsection{Results on MegaFace and Trillion-Pairs}
\begin{figure}[t]
\centering
\includegraphics[width=0.495\columnwidth]{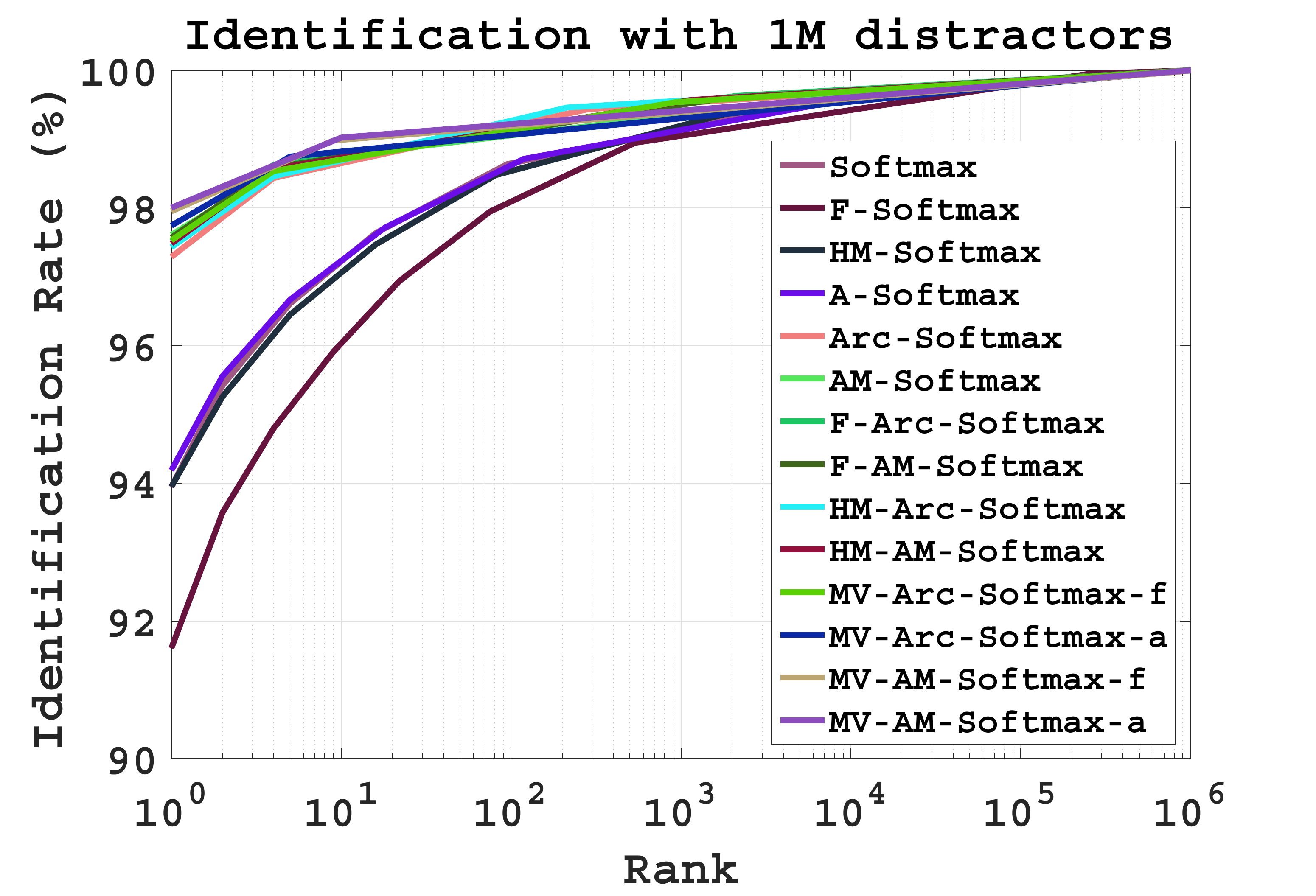}
\includegraphics[width=0.495\columnwidth]{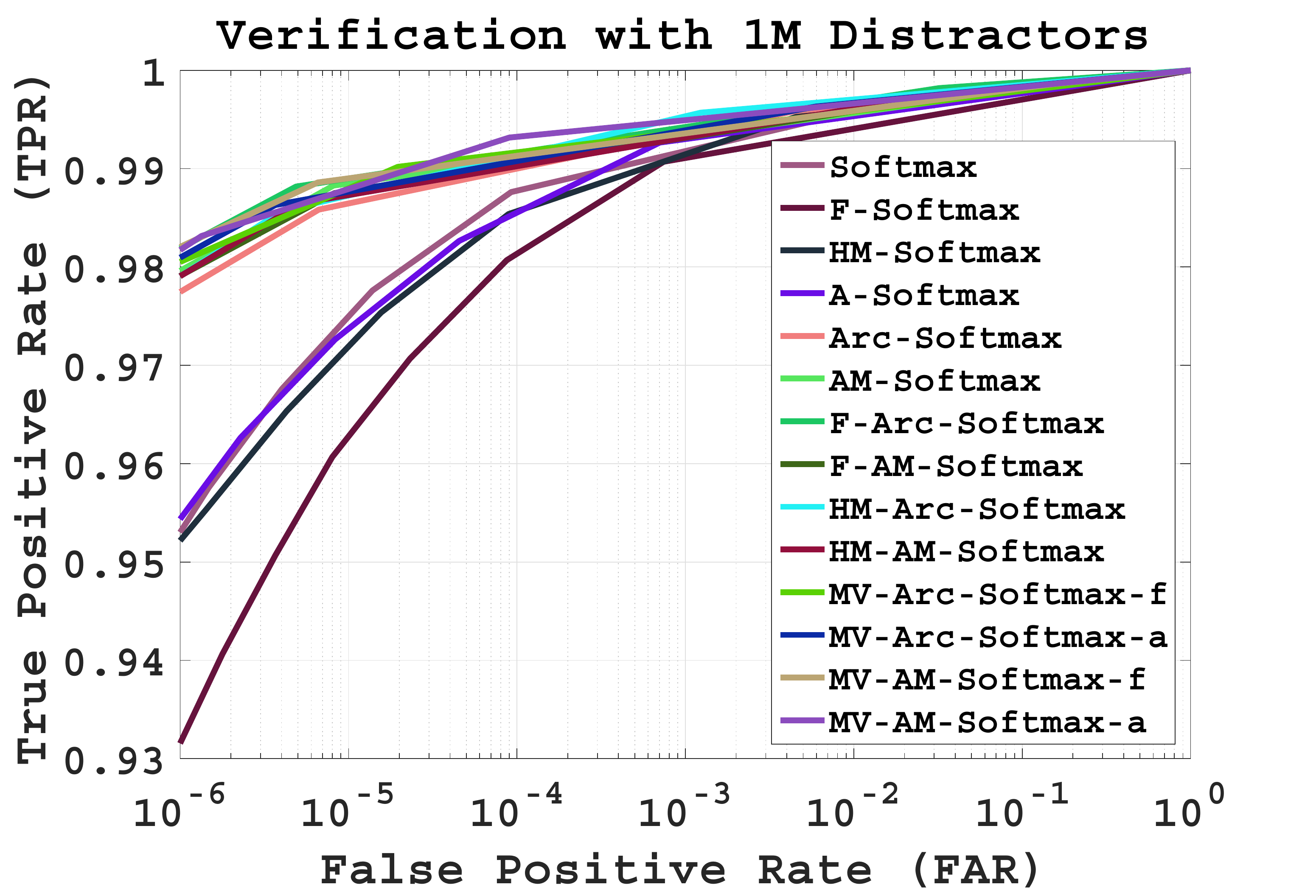}
 \caption{\textbf{From Left to Right}: CMC curves and ROC curves of different loss functions with 1M distractors on MegaFace Set 1. }
\label{fig:mega}
\end{figure}

We then test all the competitors with more complicated protocols. Specifically, the identification (Id.) Rank-1 and the verification (Veri.) TPR@FAR=1e-6 on MegaFace, the identification (Id.) TPR@FAR=1e-3 and the verification (Veri.) TPR@FAR=1e-9 on Trillion-Pairs are reported in Table \ref{MegaFace}. From the numbers, we can observe that our MV-AM-Softmax-a achieves the best performance over the baseline Softmax loss, the mining-based Softmax losses, the margin-based softmax losses and the naive combinations of mining-based and margin-based losses, on both MegaFace and Trillion-Pairs Challenge. Specifically, on MegaFace, for our proposed MV-AM-Softmax-a, it obviously beats the best margin-based competitor AM-Softmax loss by a large margin (about 0.3\% on identification and 0.5\% on verification). Compared with the naive fusions of mining-based and margin-based losses, our improved MV-AM-Softmax-a loss is also better than them. Moreover, compared the MV-Softmax-a with MV-Softmax-f, we can say that the adaptive re-weighted function Eq. (\ref{indicator2}) is generally better than the fixed one Eq. (\ref{indicator1}). This is reasonable because for more difficult mis-classfied feature vectors, they should be more important for discriminative feature learning. In Figure \ref{fig:mega}, we also draw both of the CMC curves to evaluate the performance of face identification and the ROC curves to evaluate the performance of face verification on MegaFace Set 1. From the curves, we can see the similar trends at other measures. On Trillion-Pairs Challenge, we can observe that the results exhibit the same trends that emerged on MegaFace test set. Besides, the trends are more obvious. In particular, we achieve at least 3\% improvements at both the identification and the verification on Trillion-Pairs Challenge. In this experiment, we have clearly demonstrated that our MV-AM-Softmax-a approach is superior for both the identification and verification tasks, especially when the false positive rate is very low. To sum up, by inheriting the advantages of both margin-based and mining-based Softmax losses, our new desined  mis-classified guided one has shown its strong generalization ability for face recognition.

\section{Conclusion}
This paper has proposed a simple yet very effective loss function, namely mis-classified vector guided softmax loss (\textit{i.e.}, MV-Softmax), for the task of face recognition. In specific, MV-Softmax loss explicitly concentrates on optimizing the mis-classified feature vectors. Thus it semantically inherits the motivations of feature margin and feature mining into a unified loss function. Consequently, it exhibits a higher performance than the baseline Softmax loss, the current mining-based losses, margin-based losses and their naive fusions. Extensive experiments on several face recognition benchmarks have validated the effectiveness of our new approach over the state-of-the-art alternatives.

\small
\bibliography{378_egbib}

\begin{thebibliography}{}

\bibitem[\protect\citeauthoryear{Chen, Deng, and Shen}{2018}]{chen2018virtual}
Chen, B.; Deng, W.; and Shen, H.
\newblock 2018.
\newblock Virtual class enhanced discriminative embedding learning.
\newblock In {\em NeurIPS}.

\bibitem[\protect\citeauthoryear{Deepglint}{2018}]{deepglint}
Deepglint.
\newblock 2018.
\newblock \url{http://trillionpairs.deepglint.com/overview}.

\bibitem[\protect\citeauthoryear{Deng \bgroup et al\mbox.\egroup
  }{2019}]{Arc-Softmax}
Deng, J.; Guo, J.; Xue, N.; and Zafeiriou, S.
\newblock 2019.
\newblock Arcface: Additive angular margin loss for deep face recognition.
\newblock In {\em CVPR}.

\bibitem[\protect\citeauthoryear{Feng \bgroup et al\mbox.\egroup
  }{2017}]{feng2017wing}
Feng, Z.-H.; Kittler, J.; Awais, M.; Huber, P.; and Wu, X.-J.
\newblock 2017.
\newblock Wing loss for robust facial landmark localisation with convolutional
  neural networks.
\newblock {\em arXiv:1711.06753}.

\bibitem[\protect\citeauthoryear{Guo \bgroup et al\mbox.\egroup
  }{2016}]{Msceleb}
Guo, Y.; Zhang, L.; Hu, Y.; He, X.; and Gao, J.
\newblock 2016.
\newblock Ms-celeb-1m: A dataset and benchmark for large-scale face
  recognition.
\newblock In {\em ECCV}.

\bibitem[\protect\citeauthoryear{He, Zhang, and Ren.}{2016}]{Resnet}
He, K.; Zhang, X.; and Ren., S.
\newblock 2016.
\newblock Deep residual learning for image recognition.
\newblock In {\em CVPR}.

\bibitem[\protect\citeauthoryear{Hu \bgroup et al\mbox.\egroup
  }{2015}]{hu2015face}
Hu, G.; Yang, Y.; Yi, D.; Kittler, J.; Christmas, W.; Li, S.~Z.; and
  Hospedales, T.
\newblock 2015.
\newblock When face recognition meets with deep learning: an evaluation of
  convolutional neural networks for face recognition.
\newblock In {\em CVPRW}.

\bibitem[\protect\citeauthoryear{Hu \bgroup et al\mbox.\egroup
  }{2017}]{hu2017frankenstein}
Hu, G.; Peng, X.; Yang, Y.; Hospedales, T.~M.; and Verbeek, J.
\newblock 2017.
\newblock Frankenstein: Learning deep face representations using small data.
\newblock {\em TIP}.

\bibitem[\protect\citeauthoryear{Huang, Ramesh, and Miller.}{2007}]{LFW}
Huang, G.; Ramesh, M.; and Miller., E.
\newblock 2007.
\newblock Labeled faces in the wild: A database for studying face recognition
  in unconstrained enviroments.
\newblock {\em Technical Report}.

\bibitem[\protect\citeauthoryear{Kemelmacher-Shlizerman \bgroup et
  al\mbox.\egroup }{2016}]{megaface_1}
Kemelmacher-Shlizerman, I.; Seitz, S.~M.; Miller, D.; and Brossard, E.
\newblock 2016.
\newblock The megaface benchmark: 1 million faces for recognition at scale.
\newblock In {\em CVPR}.

\bibitem[\protect\citeauthoryear{Liang \bgroup et al\mbox.\egroup
  }{2017}]{SM-Softmax}
Liang, X.; Wang, X.; Lei, Z.; Liao, S.; and Li., S.
\newblock 2017.
\newblock Soft-margin softmax for deep classification.
\newblock In {\em ICONIP}.

\bibitem[\protect\citeauthoryear{Liao \bgroup et al\mbox.\egroup
  }{2014}]{blufr}
Liao, S.; Lei, Z.; Yi, D.; and Li, S.~Z.
\newblock 2014.
\newblock A benchmark study of large-scale unconstrained face recognition.
\newblock In {\em ICB}.

\bibitem[\protect\citeauthoryear{Lin, Goyal, and Girshick.}{2017}]{Focal}
Lin, Y.; Goyal, P.; and Girshick., R.
\newblock 2017.
\newblock Focal loss for dense object detection.
\newblock In {\em ICCV}.

\bibitem[\protect\citeauthoryear{Liu \bgroup et al\mbox.\egroup
  }{2017}]{SphereFace}
Liu, W.; Wen, Y.; Yu, Z.; Li, M.; and Song., L.
\newblock 2017.
\newblock Sphereface: Deep hypersphere embedding for face recognition.
\newblock In {\em CVPR}.

\bibitem[\protect\citeauthoryear{Liu \bgroup et al\mbox.\egroup
  }{2019a}]{liu2019adaptiveface}
Liu, H.; Zhu, X.; Lei, Z.; and Li, S.~Z.
\newblock 2019a.
\newblock Adaptiveface: Adaptive margin and sampling for face recognition.
\newblock In {\em CVPR}.

\bibitem[\protect\citeauthoryear{Liu \bgroup et al\mbox.\egroup
  }{2019b}]{liu2019high}
Liu, Y.; Shi, H.; Si, Y.; Shen, H.; Wang, X.; and Mei, T.
\newblock 2019b.
\newblock A high-efficiency framework for constructing large-scale face parsing
  benchmark.
\newblock {\em arXiv preprint arXiv:1905.04830}.

\bibitem[\protect\citeauthoryear{Liu, Hu, and Wang}{2019}]{liu2019learning}
Liu, Z.; Hu, G.; and Wang, J.
\newblock 2019.
\newblock Learning discriminative and complementary patches for face
  recognition.
\newblock In {\em FG}.

\bibitem[\protect\citeauthoryear{Liu, Wen, and Yu}{2016}]{L-softmax}
Liu, W.; Wen, Y.; and Yu, Z.
\newblock 2016.
\newblock Large-margin softmax loss for convolutional neural networks.
\newblock In {\em ICML}.

\bibitem[\protect\citeauthoryear{Moschoglou \bgroup et al\mbox.\egroup
  }{2017}]{AgeDB}
Moschoglou, S.; Papaioannou, A.; Sagonas, C.; Deng, J.; Kotsia, I.; and
  Zafeiriou, S.
\newblock 2017.
\newblock Agedb: the first manually collected, in-the-wild age database.
\newblock In {\em CVPRW}.

\bibitem[\protect\citeauthoryear{Nech and Kemelmacher}{2017}]{megaface_2}
Nech, A., and Kemelmacher, I.
\newblock 2017.
\newblock Level playing field for million scale face recognition.
\newblock In {\em CVPR}.

\bibitem[\protect\citeauthoryear{Ranjan, Castillo, and
  Chellappa.}{2017}]{L2Constrain}
Ranjan, R.; Castillo, C.; and Chellappa., R.
\newblock 2017.
\newblock L2-constrained softmax loss for discriminative face verification.
\newblock {\em arXiv preprint arXiv:1703.09507.}

\bibitem[\protect\citeauthoryear{Schroff, Kalenichenko, and
  Philbin.}{2015}]{Facenet}
Schroff, F.; Kalenichenko, D.; and Philbin., J.
\newblock 2015.
\newblock Facenet: A unified embedding for face recognition and clustering.
\newblock In {\em CVPR}.

\bibitem[\protect\citeauthoryear{Sengupta \bgroup et al\mbox.\egroup
  }{2016}]{CFP}
Sengupta, S.; Chen, J.-C.; Castillo, C.; Patel, V.~M.; Chellappa, R.; and
  Jacobs, D.~W.
\newblock 2016.
\newblock Frontal to profile face verification in the wild.
\newblock In {\em WACV}.

\bibitem[\protect\citeauthoryear{Shi \bgroup et al\mbox.\egroup
  }{2017}]{shi2017cross}
Shi, H.; Wang, X.; Yi, D.; Lei, Z.; Zhu, X.; and Li, S.~Z.
\newblock 2017.
\newblock Cross-modality face recognition via heterogeneous joint bayesian.
\newblock {\em SPL}.

\bibitem[\protect\citeauthoryear{Shrivastava, Gupta, and
  Girshick.}{2016}]{OHEM}
Shrivastava, A.; Gupta, A.; and Girshick., R.
\newblock 2016.
\newblock Training region-based object detectors with online hard example
  mining.
\newblock In {\em CVPR}.

\bibitem[\protect\citeauthoryear{Simonyan and Andrew}{2014}]{VGG}
Simonyan, K., and Andrew, Z.
\newblock 2014.
\newblock Very deep convolutional networks for large-scale image recognition.
\newblock {\em arXiv preprint arXiv:1409.1556}.

\bibitem[\protect\citeauthoryear{Sun, Wang, and Tang.}{2014}]{Contrastive}
Sun, Y.; Wang, X.; and Tang., X.
\newblock 2014.
\newblock Deep learning face representation from predicting 10,000 classes.
\newblock In {\em CVPR}.

\bibitem[\protect\citeauthoryear{Sun, Wang, and Tang.}{2015}]{DeepID2+}
Sun, Y.; Wang, X.; and Tang., X.
\newblock 2015.
\newblock Deeply learned face representations are sparse, selective, and
  robust.
\newblock In {\em CVPR}.

\bibitem[\protect\citeauthoryear{Taigman, Yang, and Ranzato.}{2014}]{Deepface}
Taigman, Y.; Yang, M.; and Ranzato., M.
\newblock 2014.
\newblock Deepface: Closing the gap to human-level performance in face
  verification.
\newblock In {\em CVPR}.

\bibitem[\protect\citeauthoryear{Wang \bgroup et al\mbox.\egroup
  }{2017a}]{NormFace}
Wang, F.; Xiang, X.; Chen, J.; and Yuille., A.
\newblock 2017a.
\newblock Normface: $ l_2 $ hypersphere embedding for face verification..
\newblock In {\em ACM MM}.

\bibitem[\protect\citeauthoryear{Wang \bgroup et al\mbox.\egroup
  }{2017b}]{Attention56}
Wang, F.; Jiang, M.; Qian, C.; Yang, S.; Li, C.; Zhang, H.; Wang, X.; and Tang,
  X.
\newblock 2017b.
\newblock Residual attention network for image classification.
\newblock {\em arXiv:1704.06904}.

\bibitem[\protect\citeauthoryear{Wang \bgroup et al\mbox.\egroup
  }{2018a}]{wang2018devil}
Wang, F.; Chen, L.; Li, C.; Huang, S.; Chen, Y.; Qian, C.; and Loy, C.~C.
\newblock 2018a.
\newblock The devil of face recognition is in the noise.
\newblock In {\em ECCV}.

\bibitem[\protect\citeauthoryear{Wang \bgroup et al\mbox.\egroup
  }{2018b}]{AM-Softmax}
Wang, F.; Cheng, J.; Liu, W.; and Liu, H.
\newblock 2018b.
\newblock Additive margin softmax for face verification.
\newblock {\em SPL}.

\bibitem[\protect\citeauthoryear{Wang \bgroup et al\mbox.\egroup
  }{2018c}]{cosface}
Wang, H.; Wang, Y.; Zhou, Z.; Ji, X.; Li, Z.; Gong, D.; Zhou, J.; and Liu, W.
\newblock 2018c.
\newblock Cosface: Large margin cosine loss for deep face recognition.
\newblock {\em arXiv preprint arXiv:1801.09414}.

\bibitem[\protect\citeauthoryear{Wang \bgroup et al\mbox.\egroup }{2018d}]{RFW}
Wang, M.; Deng, W.; Hu, J.; Peng, J.; Tao, X.; and Huang, Y.
\newblock 2018d.
\newblock Racial faces in-the-wild: Reducing racial bias by deep unsupervised
  domain adaptation.
\newblock {\em arXiv:1812.00194}.

\bibitem[\protect\citeauthoryear{Wang \bgroup et al\mbox.\egroup
  }{2018e}]{wang2018support}
Wang, X.; Wang, S.; Zhang, S.; Fu, T.; and Mei, T.
\newblock 2018e.
\newblock Support vector guided softmax loss for face recognition.
\newblock {\em arXiv preprint arXiv:1812.11317}.

\bibitem[\protect\citeauthoryear{Wang \bgroup et al\mbox.\egroup
  }{2018f}]{EM-Softmax}
Wang, X.; Zhang, S.; Lei, Z.; Liu, S.; Guo, X.; and Li, S.~Z.
\newblock 2018f.
\newblock Ensemble soft-margin softmax loss for image classification.
\newblock {\em arXiv preprint arXiv:1805.03922}.

\bibitem[\protect\citeauthoryear{Wang \bgroup et al\mbox.\egroup
  }{2019}]{wang2019co}
Wang, X.; Wang, S.; Wang, J.; Shi, H.; and Mei, T.
\newblock 2019.
\newblock Co-mining: Deep face recognition with noisy labels.
\newblock In {\em ICCV}.

\bibitem[\protect\citeauthoryear{Wang, Guo, and Li}{2015}]{USSDL}
Wang, X.; Guo, X.; and Li, S.~Z.
\newblock 2015.
\newblock Adaptively unified semi-supervised dictionary learning with active
  points.
\newblock In {\em ICCV}.

\bibitem[\protect\citeauthoryear{Wang, Zhou, and Wen.}{2017}]{Angular}
Wang, J.; Zhou, F.; and Wen., S.
\newblock 2017.
\newblock Deep metric learning with angular loss.
\newblock In {\em ICCV}.

\bibitem[\protect\citeauthoryear{Wen, Zhang, and Li}{2016}]{Center}
Wen, Y.; Zhang, K.; and Li, Z.
\newblock 2016.
\newblock A discriminative feature learning approach for deep face recognition.
\newblock In {\em ECCV}.

\bibitem[\protect\citeauthoryear{Zhang \bgroup et al\mbox.\egroup
  }{2017}]{facebox}
Zhang, S.; Zhu, X.; Lei, Z.; Shi, H.; Wang, X.; and Li, S.~Z.
\newblock 2017.
\newblock Faceboxes: A cpu real-time face detector with high accuracy.
\newblock In {\em IJCB}.

\bibitem[\protect\citeauthoryear{Zhang \bgroup et al\mbox.\egroup
  }{2019}]{zhang2019faceboxes}
Zhang, S.; Wang, X.; Lei, Z.; and Li, S.~Z.
\newblock 2019.
\newblock Faceboxes: A cpu real-time and accurate unconstrained face detector.
\newblock {\em Neurocomputing}.

\bibitem[\protect\citeauthoryear{Zheng \bgroup et al\mbox.\egroup
  }{2017}]{CALFW}
Zheng, T.; Deng, W.; Hu, J.; and Hu, J.
\newblock 2017.
\newblock Cross-age lfw: A database for studying cross-age face recognition in
  unconstrained environments.
\newblock {\em arXiv:1708.08197}.

\bibitem[\protect\citeauthoryear{Zheng \bgroup et al\mbox.\egroup
  }{2018}]{CPLFW}
Zheng, T.; Deng, W.; Zheng, T.; and Deng, W.
\newblock 2018.
\newblock Cross-pose lfw: A database for studying crosspose face recognition in
  unconstrained environments.
\newblock {\em Tech. Rep}.

\bibitem[\protect\citeauthoryear{Zheng, Pal, and
  Savvides}{2018}]{zheng2018ring}
Zheng, Y.; Pal, D.~K.; and Savvides, M.
\newblock 2018.
\newblock Ring loss: Convex feature normalization for face recognition.
\newblock In {\em CVPR}.

\end{thebibliography}
\bibliographystyle{aaai}

\end{document}